\documentclass{article}

\usepackage{arxiv}

\usepackage[utf8]{inputenc} 
\usepackage[T1]{fontenc}    
\usepackage{hyperref}       
\usepackage{url}            
\usepackage{booktabs}       
\usepackage{amsfonts}       
\usepackage{nicefrac}       
\usepackage{microtype}      
\usepackage{lipsum}
\usepackage{graphicx}
\graphicspath{ {./images/} }
\usepackage[most]{tcolorbox}
\usepackage{xcolor}
\usepackage{fvextra}

\newtcolorbox{promptbox}[1]{%
  enhanced,
  breakable,
  width=\linewidth,
  before=\par\noindent,
  after=\par,
  colback=gray!8,
  colframe=blue!60!black,
  boxrule=0.8pt,
  arc=2mm,
  left=3mm,
  right=3mm,
  top=2mm,
  bottom=2mm,
  fonttitle=\bfseries,
  colbacktitle=blue!60!black,
  coltitle=white,
  title=#1,
  attach boxed title to top left={xshift=0mm,yshift=-2mm},
  boxed title style={arc=2mm,boxrule=0pt},
}

\DefineVerbatimEnvironment{PromptVerbatim}{Verbatim}{%
  fontsize=\small,
  breaklines=true,
  breakanywhere=true,
  breaksymbolleft=,
}

\title{A Retrieval-Augmented Language Assistant for Unmanned Aircraft Safety Assessment and Regulatory Compliance}

\author{
 Gabriele Immordino \\
  School of Engineering\\
  Zurich University of Applied Sciences ZHAW\\
  Winterthur, Switzerland \\
  \texttt{immo@zhaw.ch} \\
  \And
 Andrea Vaiuso \\
  School of Engineering\\
  Zurich University of Applied Sciences ZHAW\\
  Winterthur, Switzerland \\
  \texttt{vaiu@zhaw.ch} \\
  \And
 Marcello Righi \\
  School of Engineering\\
  Zurich University of Applied Sciences ZHAW\\
  Winterthur, Switzerland \\
  \texttt{rigm@zhaw.ch} \\
}

\begin{document}
\maketitle
\begin{abstract}

This paper presents the design and validation of a retrieval-based assistant that supports safety assessment, certification activities, and regulatory compliance for unmanned aircraft systems. The work is motivated by the growing complexity of drone operations and the increasing effort required by applicants and aviation authorities to apply established assessment frameworks, including the Specific Operations Risk Assessment and the Pre-defined Risk Assessment, in a consistent and efficient manner. The proposed approach uses a controlled text-based architecture that relies exclusively on authoritative regulatory sources. To enable traceable and auditable outputs, the assistant grounds each response in retrieved passages and enforces citation-driven generation. System-level controls address common failure modes of generative models, including fabricated statements, unsupported inferences, and unclear provenance, by separating evidence storage from language generation and by adopting conservative behavior when supporting documentation is insufficient. The assistant is intentionally limited to decision support; it does not replace expert judgment and it does not make autonomous determinations. Instead, it accelerates context-specific information retrieval and synthesis to improve document preparation and review while preserving human responsibility for critical conclusions. The architecture is implemented using established open-source components, and key choices in retrieval strategy, interaction constraints, and response policies are evaluated for suitability in safety-sensitive regulatory environments. The paper provides technical and operational guidance for integrating retrieval-based assistants into aviation oversight workflows while maintaining accountability, traceability, and regulatory compliance.

\end{abstract}

\section{Introduction}

The rapid growth in the number, scale, and diversity of unmanned aircraft operations is increasing the workload for aviation authorities and other stakeholders, while expectations for rigor, consistency, and traceability in safety oversight remain unchanged \cite{EASA2024EasyAccessRulesUAS}. As operational concepts evolve and the volume of regulatory material expands, assessors must spend substantial effort locating relevant passages, cross checking requirements across interconnected documents, and assembling evidence before they can form and document a justified conclusion. This effort is amplified in practice because safety and authorization workflows often span multiple artifacts, including normative rules, acceptable means of compliance, guidance, and national interpretations, where omissions or misapplied conditions can materially affect the credibility of an assessment narrative. For authorities and applicants alike, there is therefore sustained interest in tools that reduce document navigation and compilation effort without weakening accountability or changing who is responsible for regulatory decisions.

Large language models (LLMs) have demonstrated strong performance across many natural language understanding and generation tasks through large scale pretraining on heterogeneous corpora \cite{myers2024foundation, niu2024large, mitchell2023debate}. Their ability to generate fluent text and abstract patterns has supported applications in summarization, question answering, information extraction, and interactive dialogue, motivating interest in knowledge intensive professional domains, including law, medicine, and safety engineering \cite{nazi2024large,freyer2024future}. Aerospace stakeholders have also explored LLM assistance for engineering and safety critical workflows, including conceptual system design support, code generation for aerospace software artifacts, and domain specific evaluation of model factuality in aerospace manufacturing knowledge \cite{krus2025augmenting, he2025aerospacecodegen, liu2025aeromanufacturingeval}. In aviation operations, recent work studies LLM based scenario generation and decision support in air traffic settings, providing further motivation for structured, safety aware use of generative models in operational workflows \cite{gould2025airtrafficgen}.

However, direct use of standalone models in regulated workflows presents persistent limitations. Model knowledge is fixed at training time, which creates factual staleness as standards and guidance evolve \cite{bommasani2022foundation}. Language models can also produce hallucinated statements that appear credible while lacking support in authoritative sources \cite{ji2023surveyhallucination}, and they typically provide no built in provenance linking a claim to a specific regulatory passage \cite{bender2021stochastic}. In safety assessment and certification contexts, where accountability requires auditable evidence trails, these properties constrain the acceptable role of purely parametric models.

Domain adaptation through fine tuning can improve terminology usage and task behavior, and has been explored for aerospace specific knowledge and evaluation settings \cite{liu2025aeromanufacturingeval}. Yet the literature documents drawbacks that are difficult to reconcile with audit driven environments. Fine tuning requires repeated access to curated datasets and nontrivial compute, incremental updates can introduce catastrophic forgetting \cite{gekhman2024does}, and embedding domain knowledge in model parameters complicates governance and rollback compared to updating an external corpus. Recent work further suggests that parameter updates can degrade existing knowledge and can increase hallucination risk when old and new information conflict \cite{zhai2023investigating, zhang2025survey}. These observations motivate approaches that preserve a strong base model for language and reasoning, while keeping factual authority external, inspectable, and updateable.

Retrieval Augmented Generation (RAG) addresses these issues by coupling a generative model with a retrieval component that provides relevant passages at inference time \cite{lewis2020rag}, and has been applied to aviation regulatory and airworthiness question answering where citation and provenance are central requirements \cite{zheng2025airworthinessrag, zhang2025survey}. Related aviation operations literature also explores RAG style assistants for maintenance and MRO settings and adjacent structured querying tasks, emphasizing grounding over technical documentation and domain specific evaluation \cite{nagy2025crossformatrag, sutanto2025aviationmrotexttosql}. Instead of relying on parametric memory alone, a retrieval system selects candidate evidence from an external document set, and generation is conditioned on this retrieved context. This design improves factual grounding, supports explicit citation, and enables knowledge updates through corpus maintenance rather than retraining. Standard pipelines segment the corpus into smaller chunks, embed them into a vector space, retrieve the most relevant chunks using lexical or semantic similarity, and then synthesize an answer from the retrieved evidence \cite{han2025ragvsgraphrag, zhang2025survey}. Even so, baseline RAG has well documented weaknesses \cite{han2025ragvsgraphrag, zhang2025survey}. Flat chunking can ignore document structure and cross references, fragmenting logically connected requirements and reducing coherence \cite{xu2024ragkg, zheng2025airworthinessrag}. Retrieval quality often degrades for questions that require multi-step reasoning, temporal validity checks, or integration across multiple sources, and retrieval errors remain a dominant failure mode in end to end systems \cite{han2025ragvsgraphrag, masoudifard2024graphrag}. In regulated settings, this is especially consequential because missing a normative passage directly translates into incomplete guidance. Application driven evaluations report that many incomplete or incorrect answers arise from missing or inadequate retrieval rather than from generation mistakes \cite{nanua2025raven}.

Recent work proposes enhancements that better capture structure and improve traceability, which is particularly relevant for aviation regulatory usage where compliance claims must be linked to specific normative passages \cite{zheng2025airworthinessrag}. Several studies show that integrating explicit structure, often through knowledge graphs, can improve retrieval accuracy and answer completeness by preserving intra document hierarchy and inter issue relationships \cite{xu2024ragkg, agarwal2025ragulating}. Graph based retrieval can also improve requirement traceability and compliance checks by retrieving structurally related content that pure vector similarity would miss \cite{masoudifard2024graphrag}. Comparative evaluations indicate complementary strengths, conventional RAG can perform well on detail oriented queries, while GraphRAG can be more effective for reasoning intensive and multi hop questions, and hybrid strategies can yield more stable performance across query types \cite{han2025ragvsgraphrag}.

Beyond retrieval, suitability for safety related use depends on generation control and explicit handling of uncertainty. Sampling based decoding can improve multi step reasoning robustness in some settings \cite{li2025powersampling}, but compliance oriented tasks often benefit from constrained decoding paired with strict grounding and explicit refusal behavior when evidence is insufficient. Uncertainty estimation is also central, since self reported confidence is often poorly correlated with correctness, while token probability based metrics can better signal potential error \cite{bentegeac2025tokenprob}. Prompt engineering offers a practical alternative to task specific fine tuning for structured extraction and controlled outputs, but it typically requires strict schemas and validation layers to ensure consistent downstream behavior \cite{chen2025prompt}. Privacy and access control requirements further shape design, since operational details and personal data may be present in certification workflows, motivating policy enforcement and controlled anonymization mechanisms \cite{wang2025llmaccess}. These considerations align with broader analyses of foundation model risks, which emphasize that downstream systems must implement their own safeguards for confidentiality, integrity, and availability \cite{bommasani2022foundation}.

A second strand of research concerns reasoning control at inference time, with aviation motivated work demonstrating the value of controllable generation and iterative refinement for complex air traffic scenarios \cite{gould2025airtrafficgen}. Chain of Thought prompting exposes intermediate reasoning steps, allowing models to decompose complex questions into ordered sub problems and improving performance on multi-step reasoning without parameter updates \cite{wei2022cot, wang2022selfconsistency}. Tree of Thought methods extend this by exploring multiple reasoning branches, evaluating partial solutions, and selecting or refining promising paths before producing a final answer \cite{yao2023tot}. In regulatory and safety assessment contexts, such structured reasoning is attractive because it can support conservative exploration of alternatives and clearer justification of conclusions, provided that the final response remains grounded in retrieved evidence and that intermediate reasoning is handled in a way that does not encourage unsupported claims.

Despite substantial progress in retrieval and controllable generation, an open applied gap remains for unmanned aircraft safety oversight: how to engineer an assistant that is simultaneously (i) bounded to an authoritative corpus, (ii) auditable through explicit evidence linking, (iii) robust under realistic query reformulations, and (iv) usable for both document centric questions and early stage structured assessment outputs, without presenting the tool as an authority. Addressing this gap requires treating evidence control as a system level property, not only a prompting.

In this work we study retrieval augmented assistants for unmanned aircraft regulatory and safety assessment workflows with an emphasis on bounded assistance and auditability. We adopt a baseline RAG architecture with hybrid dense and lexical retrieval, conservative reranking, and explicit evidence handling, because it offers a simpler and more deterministic evidence path that is easier to validate and audit in safety sensitive settings. While graph structured retrieval remains a promising direction for multi hop queries, it introduces additional design choices around schema construction, entity and relation extraction, and traversal policies that warrant dedicated evaluation before being relied upon for regulatory decision support. The assistant is therefore treated as a controlled interface to an authoritative document set rather than an authority itself, prioritizing traceability to original sources, conservative handling of ambiguity, and explicit communication of evidence gaps, and it aligns with practical regulatory oversight constraints where the system must support careful human review rather than automate approvals or issue binding interpretations. The contributions integrate modern retrieval and reasoning techniques with explicit provenance, controlled generation, and audit-support mechanisms such as evidence-attributed explanations \cite{bosco2025ragxai}.

The remainder of the paper is structured as follows. Section~\ref{sec:methodology} describes the end-to-end methodology, including document engineering, hybrid retrieval, ranking and filtering, and controlled answer generation. Section~\ref{sec:use_cases} presents the two reference use cases that operationalize requirements for grounding and auditability in aviation and UAV workflows. Finally, Section~\ref{sec:results} reports results and observations, and Section~\ref{sec:conclusions} concludes with limitations and future directions.

\section{Methodology}
\label{sec:methodology}

This section details the methodology used to build and evaluate a retrieval-augmented assistant for unmanned aircraft system safety assessment and compliance support. It specifies how regulatory sources are ingested, represented, retrieved, ranked, and provided to the language model under explicit constraints, with the design objective of enabling traceable, auditable, and reproducible outputs suitable for human review.

At a high level, the system follows a RAG architecture with a strict separation between (i) an inspectable evidence store (the regulatory corpus and its index) and (ii) the language model used for synthesis. Regulatory documents are ingested and segmented into semantically and legally coherent chunks, enriched with metadata, and indexed with both dense (embedding-based) and sparse (lexical) retrieval backends. For each query, the system retrieves candidates, fuses dense and sparse rankings, reduces redundancy, and applies controlled reranking to produce a compact evidence set that can be logged and reviewed. The language model is invoked only after retrieval and is constrained to generate outputs strictly from the retrieved evidence under fixed decoding settings. The implementation supports two interaction modes: (i) a grounded conversational assistant for regulatory questions and (ii) an indicator-oriented workflow that returns machine-readable outputs from predefined operational parameters. The remainder of this section describes each stage in execution order.

\subsection{Data Sources and Document Engineering}
\label{subsec:data_sources}

The primary regulatory source used in this work is the Certification Specifications for Unmanned Aircraft Systems (CS-UAS) published by the European Union Aviation Safety Agency (EASA) \cite{EASA2024EasyAccessRulesUAS}. From this corpus, we selected the subset of material directly relevant to operational risk assessment, with emphasis on the normative description of the Specific Operational Risk Assessment (SORA) and Pre-defined Risk Assessment (PDRA) processes and associated guidance. The extracted subset retains definitions, stepwise assessment logic, classification criteria, and mitigation objectives, together with supporting explanatory text needed to interpret the normative requirements, while excluding unrelated certification content to keep the knowledge base intentionally scoped.

Architecturally, the pipeline is source-agnostic: extending coverage to additional regulatory documents requires only adding the new sources to the corpus and rebuilding the index, without modifying the retrieval or generation components.

\subsection{Document Preprocessing and Structural Analysis}
\label{subsec:preprocessing}

Document preprocessing is treated as a safety-relevant stage because regulatory texts impose strict requirements on semantic precision, context preservation, and provenance. Rather than extracting a flat text stream, the pipeline prioritizes structural interpretation of the source document. During parsing, visual and textual markers (e.g., headings, paragraph boundaries, lists, tables, figures, and footnotes) are detected early and represented explicitly. When possible, structure is captured directly from the parse output instead of being reconstructed post hoc from plain text, so that chunk boundaries follow the internal hierarchy of the source. This reduces ambiguity in evidence attribution and improves retrieval precision by aligning chunks to regulatory units such as articles, acceptable means of compliance (AMC), and general material entries.

Chunking is performed manually for the core regulatory content. In particular, a dedicated chunk is created for each article, AMC, and general material entry. Although this increases upfront effort, it avoids boundary ambiguities common in automated segmentation and reduces the risk of fragmenting or merging legally meaningful units. Manual chunking also simplifies provenance control: each chunk can be traced to a specific regulatory element and associated page range in the source document.

\begin{figure}[!h]
    \centering
    \includegraphics[width=1\linewidth]{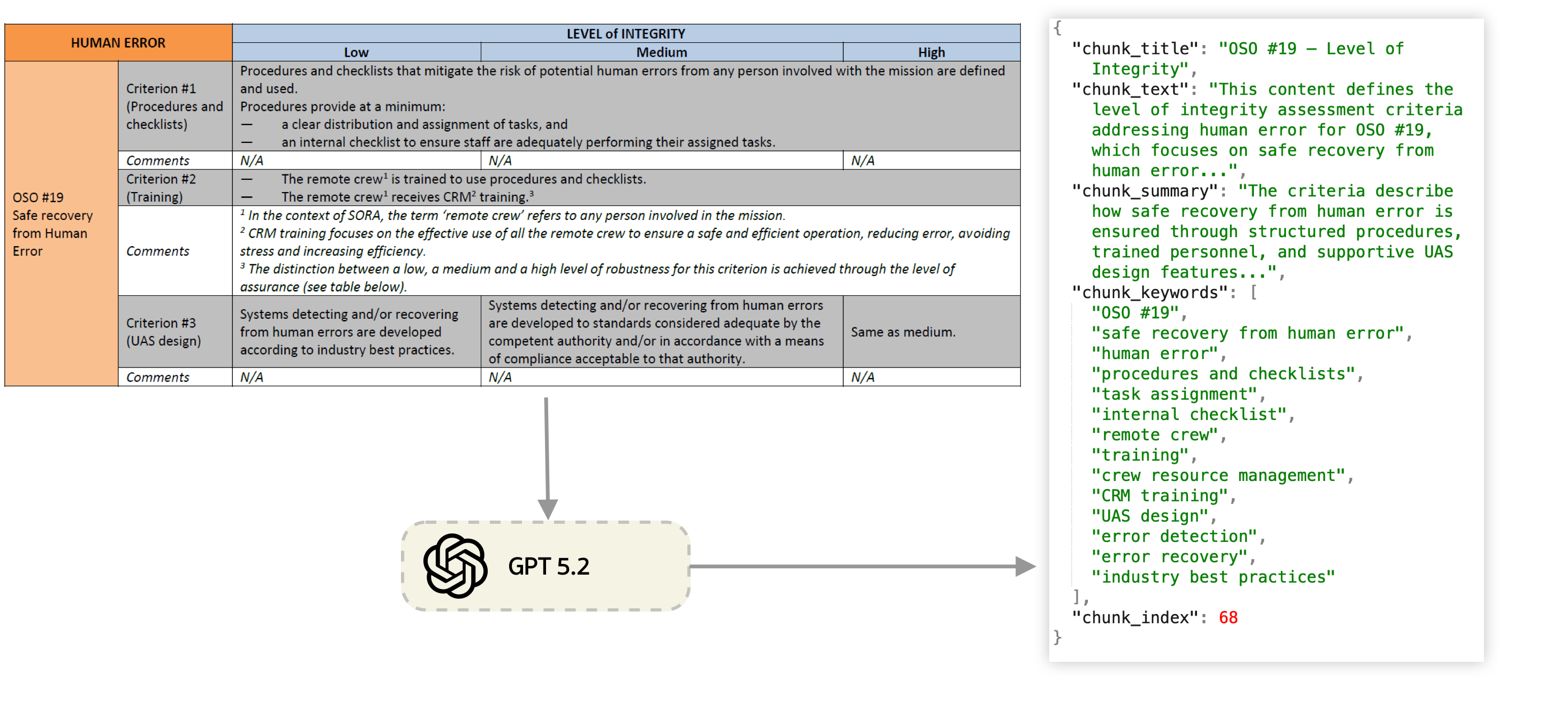}
    \caption{Table extraction process. The table is rendered as an image using ChatGPT~5.2 and converted into a JSON-formatted chunk.}
    \label{fig:table_extraction}
\end{figure}

Tables are handled via a dedicated workflow. Generic table-to-text extraction often fails to preserve cell-level associations with page-level footnotes and references; if these links are lost or mis-assigned, the resulting text can become internally inconsistent. For this reason, each extracted table is post-processed with a language model (ChatGPT~5.2) using the prompt in Appendix~\ref{appx:tab-extr} to generate (i) a canonical title, (ii) a narrative explanation that does not depend on layout while preserving the content of the cells without semantic changes, (iii) a short summary, and (iv) representative keywords. Tables are then manually chunked, assigned global identifiers, and explicitly linked to surrounding sections and paragraphs. This preserves traceability and enables later inspection of how tabular evidence influences retrieval and downstream synthesis. Figure~\ref{fig:table_extraction} illustrates the table extraction workflow.

Figures are excluded during preprocessing. In the targeted regulatory material, figures rarely contain machine-interpretable normative requirements and typically depend on visual context that is difficult to reconstruct reliably in a text-only evidence store.

\subsection{Knowledge Representation and Chunk Design}
\label{subsec:chunk_design}

After structural segmentation, each chunk becomes the fundamental unit for retrieval and evidence attribution. Each chunk is enriched with metadata fields including \texttt{chunk\_title}, \texttt{chunk\_summary}, and \texttt{chunk\_keywords} (an example is shown in Figure \ref{fig:table_extraction}). The design assigns distinct roles to each field. Titles receive higher weight during first stage retrieval, summaries provide an additional semantic signal during ranking refinement, and keywords offer a limited lexical boost. Keyword influence is intentionally constrained so that semantic similarity remains dominant. Conceptually, the keyword component behaves like a low-gain bias term in the scoring function: it rewards explicit term alignment when present, but its scale is deliberately limited relative to semantic signals. This calibration ensures that superficial word overlap cannot overrule genuine semantic relevance, while still allowing precise terminology matches to provide modest improvements in ranking fidelity.

For indexing purposes, a dedicated retrieval document is constructed by combining a title weighted representation with the main chunk text. This combined representation is generated during index construction and stored explicitly, ensuring consistent behavior across retrieval runs and supporting later inspection and audit. A further explored strategy normalizes similarity scores to a zero to one range and applies a threshold to discard weakly related evidence before generation.

\subsection{Embedding and Indexing Infrastructure}
\label{subsec:embedding_indexing}

Let $c_i$ denote the $i$-th chunk in the corpus (the textual content together with selected metadata fields used for retrieval). Each chunk is mapped to a fixed-dimensional embedding vector using a sentence-transformer encoder:
\begin{equation}
  \mathbf{v}_i = \mathrm{enc}(c_i) \in \mathbb{R}^d,
\end{equation}
where $d$ is the embedding dimension.

In our implementation, $\mathrm{enc}(\cdot)$ is instantiated with \texttt{all-MiniLM-L6-v2}~\cite{reimers2019sentencebert,wang2020minilm}. Given an input string, the model (i) tokenizes the text, (ii) encodes it with the MiniLM transformer, and (iii) produces a single sentence embedding using the standard pooling strategy provided by the sentence-transformers implementation (mean pooling over token embeddings).

The same encoder is used for both corpus items and user queries. For each chunk we embed the indexed representation $r_i$ (defined below); for a user query string $q$ we compute $\mathbf{q}=\mathrm{enc}(q)$ and its normalized form $\hat{\mathbf{q}}=\mathbf{q}/\lVert\mathbf{q}\rVert_2$.

Embeddings are computed in batches and stored as \texttt{float32} vectors. We apply L2 normalization:
\begin{equation}
  \hat{\mathbf{v}}_i = \frac{\mathbf{v}_i}{\lVert \mathbf{v}_i \rVert_2},
\end{equation}
so that the inner product between normalized vectors equals cosine similarity, i.e., $\hat{\mathbf{v}}_i^\top \hat{\mathbf{v}}_j = \cos(\theta_{ij})$. Consequently, dense retrieval can be implemented as maximum inner-product search (MIPS) over $\{\hat{\mathbf{v}}_i\}$ while retaining an interpretable similarity score in $[-1,1]$.

To improve matching when user queries reference formal section titles, we do not embed the raw chunk text directly. Instead, we construct an indexed retrieval string $r_i$ by concatenating the title and the main chunk text, repeating the title twice to up-weight its contribution in the embedding:
\begin{equation}
  r_i = \texttt{title}_i \;\Vert\; \texttt{title}_i \;\Vert\; \texttt{text}_i.
\end{equation}
Dense retrieval embeddings are computed as $\hat{\mathbf{v}}_i = \mathrm{enc}(r_i)/\lVert\mathrm{enc}(r_i)\rVert_2$. When enabled, the exact string $r_i$ is stored alongside the chunk metadata to make the indexed representation auditable.

Let $V = \{\hat{\mathbf{v}}_i\}_{i=1}^N$ be the set of normalized vectors for all $N$ indexed chunks. Given a normalized query embedding $\hat{\mathbf{q}}$, dense retrieval returns the indices of the $k$ most similar vectors:
\begin{equation}
  \mathrm{NN}_k(\hat{\mathbf{q}}) = \underset{S\subseteq\{1,\dots,N\},\ |S|=k}{\operatorname{arg\,max}}\ \sum_{i\in S} \hat{\mathbf{q}}^\top \hat{\mathbf{v}}_i.
\end{equation}
A vector index is the data structure that supports this operation efficiently.

In this work, vector indexing and search are implemented with FAISS (Facebook AI Similarity Search) \cite{johnson2017faiss}. We persist (i) the vector index itself (stored vectors plus search structure) and (ii) a separate JSON file containing chunk metadata, so that every returned vector ID can be deterministically mapped back to its source text and document location.

Two FAISS backends are supported. (i) \emph{Exact search}: a flat inner-product index that computes $\hat{\mathbf{q}}^\top \hat{\mathbf{v}}_i$ for all $i\in\{1,\dots,N\}$ and returns the top-$k$ results. Because it evaluates all candidates, this backend is deterministic and serves as the baseline for benchmarking and safety analysis when recall and auditability are prioritized over latency.

(ii) \emph{Approximate search}: a Hierarchical Navigable Small World (HNSW) index \cite{malkov2018hnsw}, which organizes the vectors as a multi-layer proximity graph $G=(\mathcal{V},\mathcal{E})$ with nodes $\mathcal{V}=\{1,\dots,N\}$ and edges linking nearby vectors. Here, each node $i\in\mathcal{V}$ corresponds to one indexed chunk embedding $\hat{\mathbf{v}}_i$, and an edge $(i,j)\in\mathcal{E}$ denotes a navigable neighbor link between two vectors selected during index construction based on their proximity in the embedding space. At query time, search starts from an entry point in the top layer and performs greedy routing toward the query, then refines the candidate set by best-first exploration in the lowest layer to approximate $\mathrm{NN}_k(\hat{\mathbf{q}})$.

HNSW exposes explicit parameters that trade off latency and recall: the maximum out-degree $M$ (here limited to 32 bidirectional links per node), the construction exploration budget \texttt{efConstruction} (here 200), and the query-time exploration budget \texttt{efSearch} (here 128). We keep these values explicit to support reproducible experiments and controlled deployment settings.

\begin{figure}
    \centering
    \includegraphics[width=1\linewidth]{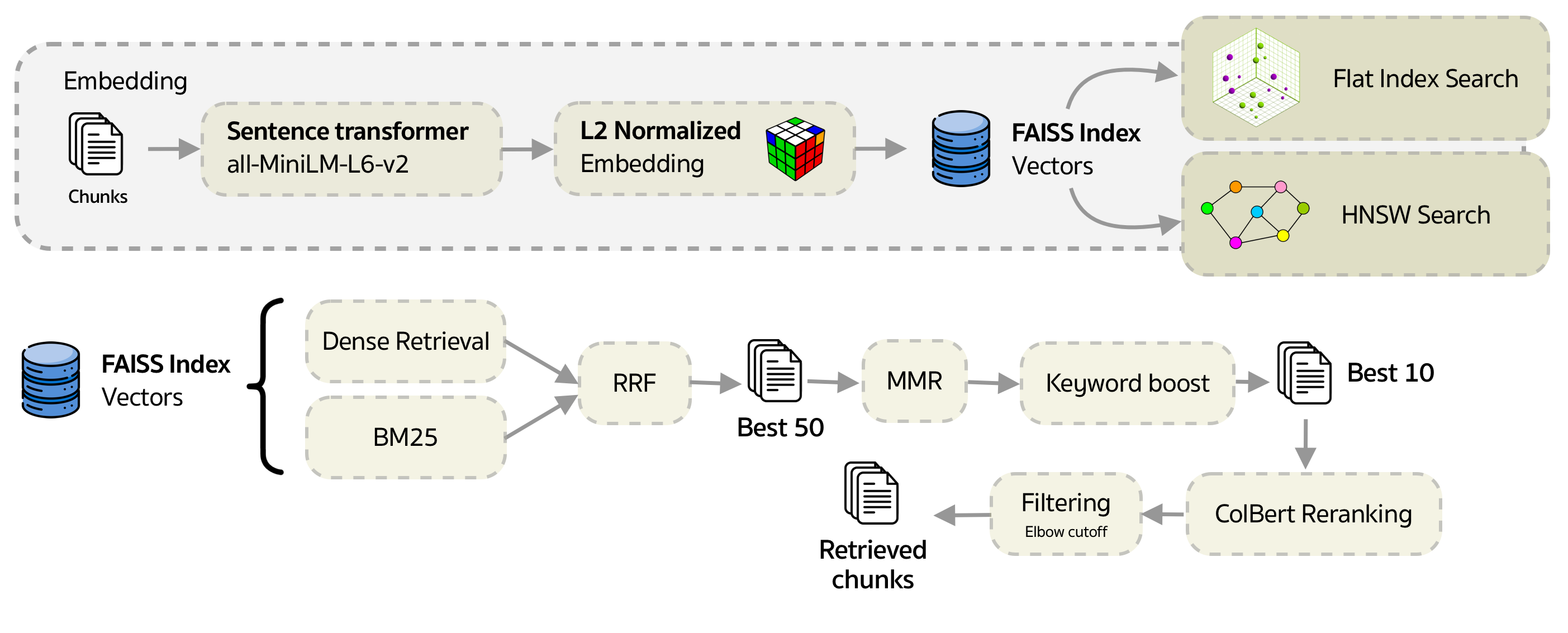}
    \caption{Embedding and retrieval architecture flowchart}
    \label{fig:rag-archit}
\end{figure}

Figure~\ref{fig:rag-archit} provides an overview of the complete retrieval and ranking pipeline, which is described in detail in the following subsections.

\subsection{Retrieval Architecture}
\label{subsec:retrieval_architecture}

The assistant follows a multi-stage retrieval and ranking pipeline designed to improve recall, reduce redundancy, and preserve auditability. Given a user query, the system performs parallel dense semantic retrieval and sparse lexical retrieval over a shared chunk representation. Dense retrieval captures conceptual relevance across paraphrases and heterogeneous regulatory language, while sparse retrieval uses BM25 (Section~\ref{subsubsec:bm25}), a probabilistic term-weighting scheme that scores documents based on query term frequency and document-length normalization \cite{robertson2009probabilistic}. The independent ranked outputs are fused using Reciprocal Rank Fusion (RRF) (Section~\ref{subsubsec:rrf}), a rank-based aggregation method that combines multiple ranked lists by summing reciprocal ranks \cite{cormack2009rrf}, yielding a hybrid candidate set that balances semantic coverage and lexical precision. The complete retrieval and ranking pipeline is summarized in Figure~\ref{fig:rag-archit}.

\subsubsection{Dense Retrieval and Diversity Control}
\label{subsubsec:dense_mmr}

Dense semantic retrieval is implemented on top of a FAISS vector index. Index vectors are normalized at build time and query embeddings are generated with normalization enabled, so inner product search corresponds to cosine similarity. Dense retrieval is oversampled to produce an initial candidate pool of 50 chunks, reducing the risk of discarding relevant material early.

Candidates are then diversified using Maximal Marginal Relevance (MMR) \cite{carbonell1998mmr}, which explicitly trades off query relevance against redundancy with already selected chunks. MMR is applied to the oversampled candidate pool to reduce near-duplicate evidence and increase coverage of distinct regulatory passages, which is important when downstream generation must be grounded in a compact context.

Operationally, MMR constructs the final list iteratively. Starting from an empty list, it repeatedly adds the candidate that is both (i) highly similar to the query (high relevance) and (ii) sufficiently different from the items already selected (low redundancy), using cosine similarity in the embedding space.

The hyperparameter $\lambda_{\mathrm{MMR}}\in[0,1]$ controls the balance between these two objectives: larger values prioritize relevance, while smaller values enforce stronger diversity. We set $\lambda_{\mathrm{MMR}}=0.6$, which favors relevance while still applying a meaningful redundancy penalty.

To support reproducibility and post hoc audit, MMR similarity computations use the stored embedding matrix when available. If embeddings are not present in memory, vectors are reconstructed from the FAISS index as a fallback.

\subsubsection{Sparse Retrieval Using BM25}
\label{subsubsec:bm25}

A sparse retrieval component based on BM25 is implemented alongside dense retrieval, operating over the same retrieval document representation. BM25 is a term-weighting function derived from probabilistic retrieval models \cite{robertson2009probabilistic}. For a query $Q$ and a document $D$, the BM25 score is computed as
\begin{equation}
\mathrm{BM25}(Q,D)=\sum_{t\in Q} \mathrm{IDF}(t)\,\frac{f(t,D)(k_1+1)}{f(t,D)+k_1\left(1-b+b\,\frac{|D|}{\overline{|D|}}\right)},
\end{equation}
where $f(t,D)$ is the frequency of term $t$ in $D$, $|D|$ is the document length, $\overline{|D|}$ is the average document length in the corpus, and $k_1$ and $b$ are hyperparameters controlling term saturation and length normalization.

Tokenization is intentionally simple and transparent. Text is lowercased and tokenized using a regular expression word extractor, avoiding stemming and aggressive normalization so that lexical matches remain directly interpretable.

\subsubsection{Hybrid Retrieval and Result Fusion}
\label{subsubsec:rrf}

The dense and BM25 ranked lists are combined using Reciprocal Rank Fusion (RRF) \cite{cormack2009rrf}, where each candidate accumulates a score of the form $\sum_i \frac{1}{k + \mathrm{rank}_i}$. The fusion constant is set to $k=60$, smoothing the influence of lower ranked items while still rewarding agreement between retrievers. The fused list is truncated to a candidate pool of 10 chunks, which is then passed to subsequent ranking refinement.

\subsection{Ranking, Filtering, and Evidence Control}
\label{subsec:ranking_filtering}

This stage takes the fused candidate pool produced by hybrid retrieval (Section~\ref{subsubsec:rrf}) and applies additional, explicitly controlled steps to (i) refine ordering using field-specific signals, (ii) apply higher-cost neural reranking, and (iii) select a small, high-quality evidence set suitable for inclusion in the bounded LLM context. The design objective is to increase precision while keeping the evidence path transparent and reproducible.

\subsubsection{Field-Aware Post-Scoring and Keywords Boosting}
\label{subsubsec:field_scoring}

After fusion, a lightweight post-scoring function refines ordering using a small number of interpretable features computed from the chunk fields. Field usage is intentionally separated across stages to avoid leakage of auxiliary fields into first-stage retrieval. Specifically, titles influence first-stage retrieval through the indexed retrieval string $r_i$ (Section~\ref{subsec:embedding_indexing}); summaries are scored only after first-stage retrieval to refine ordering within the candidate pool; and keywords are used only for a small lexical boost and are not embedded or used as the primary retrieval document.

The summary semantic score is computed by embedding \texttt{chunk\_summary} with the same encoder used for dense retrieval and measuring cosine similarity to the query embedding. The keyword boost counts how many predefined keywords appear as substrings in the lowercased query. This boost is added with a low weight so that explicit term overlap can help break ties but cannot dominate semantic relevance. The final post-score is a weighted blend of these components, with weights fixed by configuration for reproducibility.

\subsubsection{Reranking Strategies}
\label{subsubsec:rerankers}

Afterwards, the system applies a neural reranker on the already constrained candidate pool, where computational cost and latency remain bounded.


A late-interaction reranker following the ColBERT design \cite{khattab2020colbert,santhanam2021colbertv2} has been adopted. In this approach, the query and the candidate chunk are encoded separately into contextualized token embeddings. Relevance is then computed using a token-level matching operator (MaxSim): for each query token embedding, the maximum similarity over all document token embeddings is selected, and these maxima are aggregated into a final score. This design retains token-level alignment signals (e.g., matching specific regulatory terms) while avoiding full cross-attention between the two sequences. A key consequence is that document token embeddings can be precomputed and reused across queries, making the method more efficient than cross-encoding as corpus size grows, while remaining more expressive than a single-vector bi-encoder.

\subsubsection{Result Filtering}
\label{subsubsec:filtering}

After post-scoring and reranking, the system selects a compact set of evidence chunks using an elbow method over the ranked score sequence. Results are sorted by score and consecutive drops are computed as $\Delta_i = s_i - s_{i+1}$. The elbow point is identified as the first index where the score drop exceeds a fixed threshold (0.8 in our implementation); the method then keeps only the top-ranked prefix up to that elbow. This approach suppresses low-score tails that frequently correspond to weakly related context, reducing the risk of diluting the evidence set used for generation.

\subsection{Large Language Model Integration and Answer Generation}
\label{subsec:llm_integration}

\begin{figure}[h]
    \centering
    \includegraphics[width=1\linewidth]{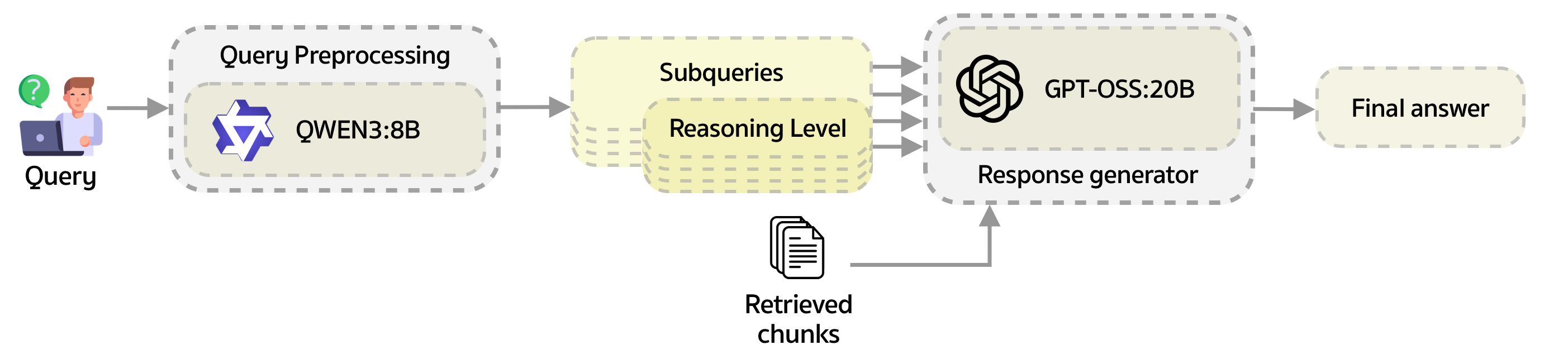}
    \caption{Response schema for grounded answers and query preprocessing}
    \label{fig:response_schema}
\end{figure}

The final stage uses a LLM strictly for synthesis and structured response generation using only the retrieved regulatory material. The model is not permitted to introduce external sources, infer intent beyond cited text, or perform independent classification steps not supported by evidence. Each retrieved chunk is accompanied by metadata such as section title, internal identifier, source document, and page reference. Figure~\ref{fig:response_schema} summarizes the response schema architecture.

Retrieved content is assembled into a single context block with a fixed maximum budget of approximately 12{,}000 characters. Chunks are inserted in ranked order until the budget is reached, ensuring deterministic context construction. Chunk identifiers are preserved inline so each statement can be traced to its originating passage. Retrieval parameters are fixed for experimental reproducibility. The system uses an initial retrieval pool of 50 candidates and retains 10 chunks for answer generation after reranking or filtering.

\subsubsection{Optional Query Preprocessing}
\label{subsubsec:query_preprocess}

When enabled, query preprocessing rewrites or decomposes an incoming user request into a small set of focused retrieval queries. The objective is to sharpen the retrieval signal for compound or underspecified questions by splitting independent sub-questions, reducing ambiguity, and generating more targeted search phrases, while keeping downstream answer generation grounded in retrieved regulatory passages.

In our configuration, this preprocessing step is performed by a dedicated query generator model based on Qwen3-8B-Instruct \cite{qwen2024qwen25}. The generator returns at most $n$ sub-queries, each paired with a \texttt{reasoning\_level} tag selected from \texttt{none}, \texttt{low}, \texttt{medium}, or \texttt{high}, following the fixed JSON format, using the prompt reported in Appendix \ref{appx:query-preprocessing}. Each sub-query is then processed sequentially through the same retrieval and generation pipeline; conversational state is reset between sub-queries to avoid unintended cross-contamination of evidence.

 

\subsubsection{Controlled Generation and Evidence Gaps}
\label{subsubsec:controlled_generation}

Answer generation is governed by fixed system and developer instructions that enforce context-only answering, citation-driven synthesis, and conservative handling of ambiguity. If the retrieved material is insufficient to support a requested conclusion, the model is instructed to explicitly state the evidence gap and avoid speculative completion.

OpenAI \texttt{gpt-oss-20B} was chosen as LLM backend \cite{openai2025introducinggptoss} to support controlled generation in a safety-oriented setting while preserving a bounded and auditable evidence path. In particular, when this backend exposes an internal reasoning trace, the system can capture it during answer construction and store it separately from the user-facing response. This trace supports debugging and expert review (e.g., to detect reasoning failures, prompt leakage, or unsupported inference patterns) and can be used to reproduce and compare runs; however, it is not treated as evidence and is never used to justify regulatory claims.

Generation parameters are fixed to support reproducibility. The maximum number of generated tokens is set to approximately 56{,}000, temperature to 0.2, top-$p$ to 0.9, and presence and frequency penalties to 0. A single completion is produced per query. A configurable \texttt{reasoning\_effort} flag is passed to the backend.

To support audit and replay, the system stores the full prompt context for each answer, including the ordered list of retrieved chunk identifiers, the bounded context block, the final list of cited sources, and when available the internal reasoning trace.

\subsection{Indicator Oriented Safety Assessment Workflow}
\label{subsec:indicator_workflow}

In addition to conversational question answering, the system supports an indicator oriented prompting workflow for structured safety assessment tasks. Each invocation requests exactly one indicator to keep outputs focused and easier to validate. The assistant is required to return exactly one JSON object following a fixed schema:
\[
\{\texttt{"name"}:\dots,\ \texttt{"value"}:\dots,\ \texttt{"explanation"}:\dots\}.
\]
The schema constraint is included directly in the request and the assistant is instructed to return no additional text, supporting deterministic parsing and reducing the risk of mixing narrative and machine readable content.

Operational parameters are captured in a structured input object where each field accepts only predefined values, including visual containment mode (VLOS or BVLOS), ground environment, airspace type, mass category, and altitude category. Prompts are minimized by including only the input fields relevant to the requested indicator, limiting ambiguity and avoiding irrelevant details.

For each indicator, supporting material is retrieved using the same retrieval pipeline. Retrieval queries are constructed deterministically from indicator specific base terms extended with controlled vocabulary derived from the selected operational categories. Retrieved chunks are assembled into a single context block with the same 12{,}000 character limit. Traceability is preserved by recording included chunk identifiers and reconstructing a human readable sources list including chunk index, originating file name, section title, and page number.


\section{Use Cases}
\label{sec:use_cases}

This work is structured around two reference use cases that define the intended scope of the system and provide concrete criteria for evaluating its behavior in regulatory and safety assessment contexts. The use cases are not conceived as production automation. They are controlled experimental scenarios that enable systematic analysis of feasibility, limitations, and design trade offs when applying language model based assistants to regulatory workflows.

Both use cases reflect certification and authorization activities for unmanned aircraft operations, where outputs must be verifiable, repeatable, and suitable for review by a competent human assessor. As mentioned before, the system is treated as decision support rather than an authority. It is intended to assist with document navigation, requirement extraction, and structured summarization, while avoiding any behavior that could be interpreted as issuing approvals, making legal determinations, or substituting expert judgement.

The use cases also serve as a concrete test of the safety safeguards implemented in the system. In both scenarios, the assistant is restricted to using only the retrieved regulatory passages. It must also identify evidence gaps rather than making assumptions, and provide traceability through explicit source references (e.g. chunk identifiers and page links) for review and audit purposes. Additionally, interaction and logging choices are designed to minimise the risk of disclosing sensitive operational information. Together, these use cases provide scenarios for evaluation, with explicit, testable criteria for grounding, traceability, conservative behaviour in situations of uncertainty, and robustness across interaction modes.

\subsection{Use Case 1, Regulatory Assistant}
\label{subsec:uc_reg_assistant}

The first use case evaluates a controlled conversational assistant grounded exclusively in official regulatory material. The objective is to demonstrate that a language model based system can operate within a bounded knowledge space without introducing external assumptions, undocumented practices, or speculative interpretations.

The assistant is constrained to rely only on authoritative sources such as SORA guidance, PDRA descriptions, and associated advisory material. The retrieval corpus is treated as the sole source of truth, and responses must not draw on general world knowledge unless it can be traced to the provided documents. This constraint targets common failure modes of language models, including hallucinated content, inconsistent reasoning across turns, and over generalization when related but non identical regulatory contexts exist.

A core requirement is traceability. The assistant is expected to justify claims by pointing to retrieved passages that support them, and to separate document grounded statements from explanatory phrasing. When retrieved context is incomplete, ambiguous, or conflicting, the assistant should explicitly signal the limitation, specify what additional information is needed, and, when appropriate, present multiple plausible interpretations consistent with the sources. If no adequate support exists, it should avoid guessing and return an uncertainty bounded response.

Evaluation focuses on factual and procedural questions, clarification of regulatory terminology, explanation of applicability and pathways, and description of documentation requirements. Typical interactions include explaining PDRA applicability, distinguishing operational categories, and summarizing steps in authorization processes. Beyond correctness, this use case tests robustness and usability, including stability across prompt length, streaming modes, retrieval settings, and follow up questions. The assistant should also manage mismatches between user intent and regulatory language, for example when informal terms are used, categories are mixed, or required parameters are not provided, such as drone class marking, operational volume characteristics, population density, airspace class, or containment assumptions. In these cases, the assistant should ask targeted clarification questions and explain why the missing parameters matter, without inventing them.

This use case serves as a baseline validation of the retrieval augmented architecture and the minimum safeguards required for safety oriented deployment considerations. The structured characteristics of this scenario, including actors, preconditions, system flow, and acceptance criteria, are summarized in Table~\ref{tab:uc_chatbot}. These safeguards include prompt rules that prohibit unsupported claims, a consistent citation or snippet mechanism, logging that enables audit without storing unnecessary personal data, and refusal behaviour for requests that fall outside the document set or exceed the role of an assistant.

\begin{table}[h]
\centering
\renewcommand{\arraystretch}{1.2}
\begin{tabular}{|p{4cm}|p{10cm}|}
\hline
\textbf{Item} & \textbf{Description} \\
\hline
Use Case ID & UC-CHAT-01 \\
\hline
Use Case Name & Ask SORA or drone operations question via chatbot \\
\hline
Primary Actor & User, operator, safety engineer, student \\
\hline
Supporting Actors or Systems & Web GUI or notebook client; RAG system, vector search and reranker; LLMChatbot; QueryGenerator (optional) \\
\hline
Goal & Get a correct, explainable answer grounded in SORA related documents, with references or snippets where possible \\
\hline
Trigger & User submits a natural language question, for example ground risk buffer, OSOs, or mitigations \\
\hline
Preconditions & RAG index is built and loaded, RAG system available; LLM credentials and configuration present; network access to model endpoint if required \\
\hline
Inputs & Question; optional chat history; retrieval parameters (top-$k$ and ce\_keep\_k); optional reasoning effort \\
\hline
Main Success Scenario &
1. User asks a question.
2. Optionally, the system expands the question into sub-queries with QueryGenerator.
3. RAG retrieves candidate chunks using top-$k$ and reranks or filters using ce\_keep\_k.
4. The system builds a grounded prompt from the question, retrieved context, and chat history if available.
5. The LLM generates an answer, optionally streamed.
6. The system returns the final response to the user. \\
\hline
Error or Exception Flows &
E1. Model endpoint unavailable or timeout, return an error with retry guidance.
E2. Missing credentials or configuration, fail fast with an actionable message.
E3. Retrieval index missing or corrupted, instruct to rebuild the index or check the processed files index. \\
\hline
Outputs & Answer text, possibly streamed; optional citations such as chunk titles or identifiers; logs or telemetry if enabled \\
\hline
Postconditions & User receives a response; the system may store the conversation turn if persistence is enabled \\
\hline
Non Functional Requirements &
Latency, response within acceptable time, streaming preferred.
Quality, grounded and consistent with sources.
Safety, avoid hallucinations and show uncertainty.
Privacy, do not leak secrets or sensitive data from logs or configuration. \\
\hline
Acceptance Criteria &
For a representative SORA question, the chatbot returns
1. a coherent answer,
2. uses retrieved context,
3. handles empty retrieval gracefully,
4. does not crash when streaming is enabled. \\
\hline
\end{tabular}
\label{tab:uc_chatbot}
\caption{Chatbot use case, retrieval augmented SORA question answering.}
\end{table}

\subsection{Use Case 2, Safety Assessment Task}
\label{subsec:uc_safety_assessment}

The second use case targets early stage analytical tasks where the assistant produces structured assessment outputs from a compact set of user provided operational parameters. The intent is to evaluate whether the assistant can support initial safety assessment work by organizing information, following standard reasoning steps, and producing outputs that are consistent in form and grounded in the same authoritative material used in the regulatory assistant use case.

Inputs represent a simplified description of an intended operation and cover parameters that commonly influence regulatory classification and risk characterization, including aircraft characteristics, operational mode, ground environment, and airspace context. Inputs are assumed to be provided by a competent user or derived from an upstream planning step. The assistant does not infer missing parameters. Inputs are expected to be filled, and the assistant should treat them as given rather than optional.

A representative task is generation of an initial SORA oriented assessment summary. Given the operational parameters, the assistant is expected to produce structured outputs such as a plausible regulatory pathway, an initial ground risk orientation, an initial air risk orientation, and an indication of expected assessment depth. The output is intended to support documentation preparation and to provide a structured starting point for a human assessor rather than to replace formal assessment activity.

The task is framed as a bounded analytical aid. The assistant should remain conservative, communicate uncertainty explicitly when evidence is insufficient, and highlight which additional parameters or documents would be required to strengthen the conclusion. Where multiple interpretations may apply, it should present alternatives and explain the conditions under which each would be appropriate, while remaining within the scope of the referenced material. Evaluation focuses on workflow adherence, internal consistency across indicator outputs derived from the same inputs, and avoidance of unsupported claims. Particular attention is given to over confidence, silent assumptions, inconsistent categorization, and omission of key decision conditions. The assistant must avoid normative phrasing that could be misread as an approval or binding compliance judgement.

The structured definition of actors, inputs, system flow, and acceptance criteria for this analytical scenario is summarized in Table~\ref{tab:uc_cls_indicators}. This table formalizes the indicator based workflow and defines the boundary conditions under which the task is considered successfully executed.

\begin{table}[h]
\centering
\renewcommand{\arraystretch}{1.2}
\begin{tabular}{|p{4cm}|p{10cm}|}
\hline
\textbf{Item} & \textbf{Description} \\
\hline
Use Case ID & UC-CLS-01 \\
\hline
Use Case Name & Classify initial operation and return indicator answers \\
\hline
Primary Actor & User, operator, analyst, student \\
\hline
Supporting Actors or Systems & Notebook or Web GUI client; RAG system, vector search and reranker; LLM based indicator assistant (LLMIndicatorAssistant) \\
\hline
Goal & Produce consistent initial indicators, for example regulatory pathway and initial ground or air risk orientation, grounded in SORA documentation \\
\hline
Trigger & User provides an InitialOperationInput and selects one or more indicators to evaluate \\
\hline
Preconditions & RAG index is built and loaded, RAG system available; LLM credentials and configuration present; LLMIndicatorAssistant initialized, engine available \\
\hline
Inputs & Operation descriptor \texttt{op} including mass category, VLOS or BVLOS, ground environment, airspace type, altitude category; indicators list; flags such as \texttt{stream} and \texttt{print\_sources} \\
\hline
Main Success Scenario &
1. User defines \texttt{op}.
2. User selects indicators to compute.
3. For each indicator, the system retrieves relevant chunks from the RAG index.
4. The system prompts the LLM to produce a structured indicator result.
5. The system aggregates results into a results dictionary.
6. The system prints results and or exports them to JSON, for example an initial indicators JSON file. \\
\hline
Error or Exception Flows &
E1. Invalid \texttt{op} values, return a validation error or ask for correction.
E2. RAG retrieval fails or returns empty context, return a low confidence output with a clarification request.
E3. Model endpoint unavailable or timeout, return an error and allow retry.
E4. JSON export path invalid, return a file write error. \\
\hline
Outputs & Results per indicator containing classification and optional rationale; optional sources or citations; optional JSON export \\
\hline
Postconditions & Results are available for downstream steps, such as later SORA workflow, reporting, or user interface display \\
\hline
Non Functional Requirements &
Consistency, similar inputs yield stable outputs.
Auditability, rationale and sources available when enabled.
Latency, acceptable runtime for multiple indicators.
Safety, avoid hallucination and clearly communicate uncertainty. \\
\hline
Acceptance Criteria &
For a valid \texttt{op} and a non empty indicators list, the system produces a results dictionary containing all requested indicator keys, prints them without crashing, and can optionally export them to JSON. \\
\hline
\end{tabular}
\label{tab:uc_cls_indicators}
\caption{Classification task use case, initial indicators.}
\end{table}

\subsection{Use Cases as Design and Evaluation References}
\label{subsec:uc_as_references}

Together, the two use cases define the system scope and support a consistent evaluation philosophy. The regulatory assistant provides a conservative reference focused on bounded sources, traceability, and predictable behavior. The safety assessment task extends the system toward structured analytical assistance, while still treating outputs as decision support subject to human review.

These use cases translate high level goals into testable expectations about inputs, outputs, and acceptable failure behavior. They inform architectural choices, motivate safeguards, and shape evaluation through criteria such as grounding, consistency, uncertainty handling, and robustness under incomplete evidence. They also provide a structured basis for discussing future extensions, such as broader coverage of assessment steps and artifacts or adaptation to other regulatory contexts, under the same principles of bounded sources, transparent reasoning support, and clear separation between assistance and authority.

\section{Results}
\label{sec:results}

This section reports results for two use cases: (i) a regulatory grounded conversational assistant and (ii) a structured safety assessment task assistant that outputs early stage SORA oriented indicators.

\subsection{Regulatory assistant}
\label{subsec:results_regulatory}

\begin{figure}[ht]
    \centering
    \includegraphics[width=\linewidth, trim=0 0 0 5.2cm, clip]{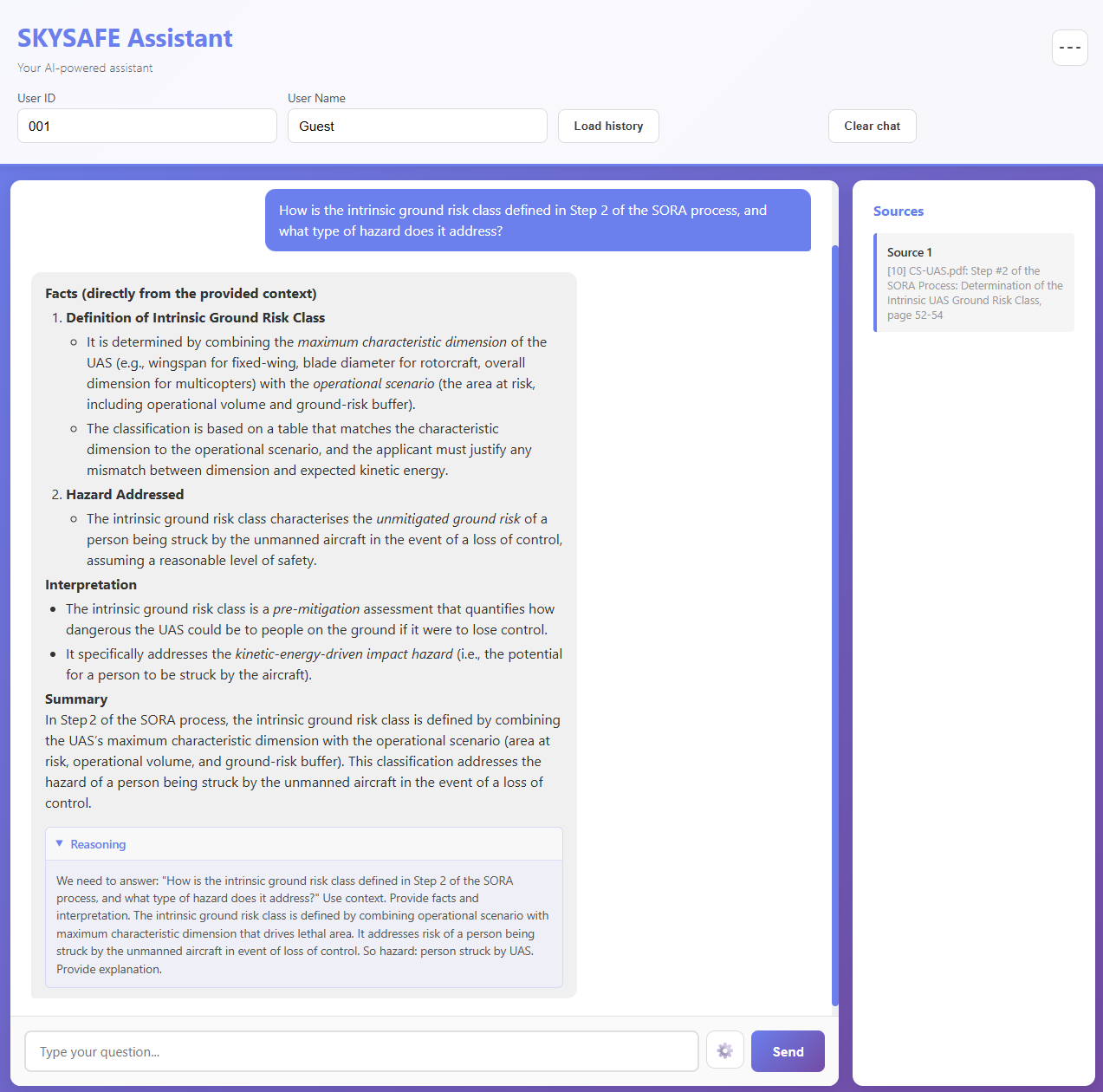}
    \caption{Interface for the regulatory assistant, showing a grounded answer with a sources panel.}
    \label{fig:ui_chat}
\end{figure}

Figure~\ref{fig:ui_chat} illustrates the interface used for the regulatory assistant. The interface exposes the user query, the generated answer structured into labeled sections, and a sources panel that lists the retrieved chunks supporting the response. This layout reflects the grounding and traceability constraints described earlier, making the evidence path visible to the reviewer.

Across typical factual and procedural questions, the assistant generally produced answers aligned with retrieved material when the governing paragraph or table excerpt was present in the retrieved context. The most reliable interactions mapped cleanly to a single definition, threshold, or procedural requirement, enabling the assistant to restate requirements while preserving traceability for assessor review.

Retrieval accuracy was evaluated via a controlled experiment with $N=5$ source chunks. For each chunk, $k=3$ query variants were constructed, yielding $|Q| = Nk = 15$ evaluation queries. Variants included a direct match formulation, a synonym paraphrase, and a reworked question with altered phrasing and conceptual framing. Retrieval quality was quantified using Hit@$m$ (for $m \in \{1,3,5,10\}$) and mean reciprocal rank (MRR).

\begin{table}[ht]
\centering
\label{tab:retrieval_metrics_variant}
\begin{tabular}{lccccc}
\hline
Variant & Hit@1 & Hit@3 & Hit@5 & Hit@10 & MRR \\
\hline
Direct match & 100.0\% & 100.0\% & 100.0\% & 100.0\% & 1.000 \\
Synonym paraphrase & 75.0\% & 75.0\% & 100.0\% & 100.0\% & 0.800 \\
Reworked question & 75.0\% & 75.0\% & 75.0\% & 75.0\% & 0.750 \\
\hline
Overall & 83.3\% & 83.3\% & 91.7\% & 91.7\% & 0.850 \\
\hline
\end{tabular}
\caption{Retrieval metrics by question variant for the five chunk accuracy test.}
\end{table}

Direct match queries produced perfect retrieval across all cutoffs, indicating that terminology aligned with the source chunk leads to stable top ranked evidence. Under synonym paraphrases, the correct chunk remained retrievable within the top 5 and top 10, but rank 1 precision degraded, consistent with semantically adjacent passages sometimes being ranked above the ground truth. Reworked questions exhibited the largest degradation, and increasing the cutoff did not recover the missing cases, consistent with a recall limitation under substantial semantic reformulation rather than only a reranking issue.

Retrieval success did not guarantee that generated answers were fully supported by the retrieved context. Grounding was assessed with a sentence level support check using the retrieved context and the citations emitted by the model. Each answer was classified as \emph{Grounded} when all sentences were supported and citations substantiated claims, \emph{Unsupported} when at least one sentence lacked support or citations failed to justify it, and \emph{Incomplete} when only partial evidence was available for the required response. In addition, \emph{Chunk Used} measures whether the generation used or cited the ground truth chunk rather than relying only on adjacent passages.

\begin{table}[ht]
\centering
\label{tab:grounding_outcomes_variant}
\begin{tabular}{lcccc}
\hline
Variant & Grounded & Unsupported & Incomplete & Chunk used \\
\hline
Direct match & 50.0\% & 0.0\% & 0.0\% & 50.0\% \\
Synonym paraphrase & 25.0\% & 25.0\% & 0.0\% & 50.0\% \\
Reworked question & 50.0\% & 25.0\% & 0.0\% & 25.0\% \\
\hline
\end{tabular}
\caption{Grounding outcomes by question variant.}
\end{table}

Two patterns stand out. First, grounded answers were less frequent than correct retrieval alone would suggest. Even under direct match queries with perfect retrieval, only half of answers were classified as grounded, and only half explicitly used the ground truth chunk, indicating gaps in citation coverage and partial exploitation of retrieved evidence during generation. Second, for reworked questions, chunk usage dropped to 25\% even when groundedness returned to 50\%, suggesting that supported answers may rely on alternative retrieved passages when the intended ground truth chunk is not retrieved reliably.

When retrieval was weak or fragmented across multiple chunks, the assistant tended to produce plausible but incomplete responses, often omitting conditions, exceptions, or parameter thresholds. Follow up questions benefited from preserved conversational context when retrieval remained anchored on the same topic, while topic drift in retrieval could lead to response drift. Observed failure patterns included degradation under weak retrieval, reduced coherence when evidence was split across chunks, and implicit assumptions triggered by ambiguous user phrasing. These findings motivate conservative behaviour when evidence is incomplete, together with retrieval tuning for higher coverage on queries that imply conditions, exceptions, or composite requirements.

\subsection{Safety assessment task}
\label{subsec:results_safety_assessment}

\begin{figure}[ht]
    \centering
    \includegraphics[width=\linewidth, trim=0 0 0 4.5cm, clip]{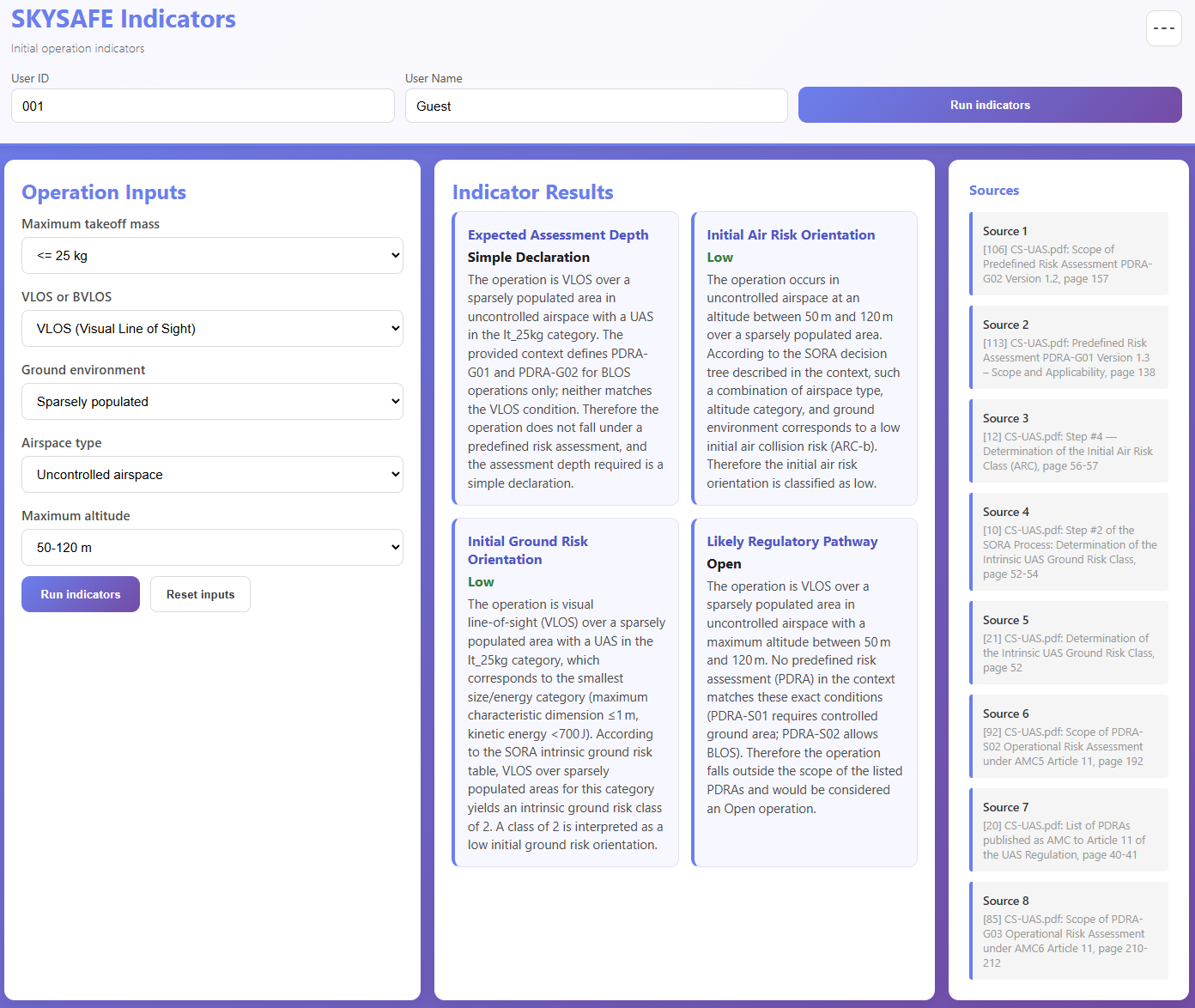}
    \caption{Indicator workflow interface, showing operation inputs, computed indicators, and supporting sources.}
    \label{fig:ui_indicators}
\end{figure}

Figure~\ref{fig:ui_indicators} shows the indicator oriented workflow interface used in the structured safety assessment task. The interface presents the categorical operation inputs, the computed indicator outputs, and the associated sources that informed each result. This layout mirrors the single indicator JSON contract described in the methodology and supports inspection of how structured inputs propagate into categorized outputs.

The safety assessment task produces four categorical indicators from structured operational inputs: likely regulatory pathway, initial ground risk orientation, initial air risk orientation, and expected assessment depth. Five representative operation configurations were evaluated. For each operation and indicator, the model was executed $N=10$ times, and a majority vote was used as the final prediction.

Consistency was assessed via (i) value consistency, measuring stability of the categorical output, and (ii) explanation similarity, measuring stability of the generated rationale text using a surface form similarity proxy. Table~\ref{tab:consistency_metrics} reports mean results aggregated across the five operations.

\begin{table}[ht]
\centering
\label{tab:consistency_metrics}
\begin{tabular}{lcc}
\hline
Indicator & Value consistency [\%] & Explanation similarity [\%] \\
\hline
Likely regulatory pathway & 83.3 & 59.3 \\
Initial ground risk orientation & 86.7 & 43.9 \\
Initial air risk orientation & 96.7 & 60.2 \\
Expected assessment depth & 100.0 & 66.9 \\
\hline
Overall & 91.7 & 57.6 \\
\hline
\end{tabular}
\caption{Mean consistency metrics per indicator across repeated runs.}
\end{table}

Categorical outputs were highly repeatable overall, with mean value consistency of 91.7\%. Expected assessment depth was fully stable, and initial air risk orientation was also highly stable, suggesting convergence to a single category under repeated execution. Likely regulatory pathway showed the lowest value stability, consistent with being more sensitive to interpretation boundaries.

Explanation similarity was consistently lower than value consistency, indicating that the assistant often converged to the same categorical decision while varying wording and structure in the rationale. The lowest explanation similarity occurred for initial ground risk orientation, indicating especially high surface level variation in ground risk justifications even when the category remained stable.

Accuracy was computed by comparing majority vote predictions to manually curated ground truth labels. Table~\ref{tab:cls_accuracy} reports per indicator and overall accuracy.

\begin{table}[ht]
\centering
\label{tab:cls_accuracy}
\begin{tabular}{lc}
\hline
Indicator & Accuracy [\%] \\
\hline
Likely regulatory pathway & 66.6 \\
Initial ground risk orientation & 80.3 \\
Initial air risk orientation & 80.3 \\
Expected assessment depth & 100.0 \\
\hline
Overall & 81.8 \\
\hline
\end{tabular}
\caption{Accuracy of majority vote predictions against ground truth.}
\end{table}

Overall accuracy reached 81.8\%. Expected assessment depth achieved 100.0\% accuracy, matching its perfect value stability. Initial ground risk orientation and initial air risk orientation both achieved 80.3\%, indicating strong agreement on coarse risk orientation assignment for the evaluated operations. Likely regulatory pathway achieved 66.6\% and remained the most challenging indicator, consistent with its lower stability and sensitivity to interpretive boundaries.

Although the inputs were complete, some decisions depended on regulatory interpretation rather than missing data. In these cases, the most reliable behaviour occurred when outputs were framed as conditional or indicative rather than selecting a single definitive classification for borderline cases. Grounding remained the dominant determinant of correctness. When retrieval returned directly relevant passages, explanations aligned closely with the cited text and avoided unsupported criteria, while partial retrieval reduced completeness and specificity.

Two practical failure modes were observed: over specificity in borderline cases and inconsistency across outputs when indicators were generated independently under differing implicit assumptions. These results motivate safeguards beyond prompt constraints, including explicit signalling of conditional outcomes, validation checks for cross indicator coherence, and conservative handling when evidence coverage is incomplete.

\section{Conclusions}
\label{sec:conclusions}

This work examined whether a retrieval augmented language model assistant can support unmanned aircraft safety assessment and regulatory compliance tasks while remaining traceable, reviewable, and compatible with safety critical expectations. The system was framed as a decision support tool with strict boundaries: authoritative documents are treated as the sole source of truth, outputs must remain evidence based, and responsibility for assessment outcomes remains with qualified human assessors and competent authority processes. 

Across the two reference use cases, three conclusions follow. First, a regulatory assistant is feasible when retrieval returns the governing paragraph or table excerpt. Under these conditions, the assistant behaves as an access interface to the corpus, helping reduce time spent on document navigation and restating requirements with explicit sources suitable for review. 

Second, retrieval quality is necessary but not sufficient. Controlled tests indicate strong performance when query terminology aligns with the source, while paraphrasing and heavily reworked questions reduce recall, and even correct retrieval can be undermined if generation does not consistently attribute statements to the most relevant evidence. This supports a design stance where evidence selection, citation discipline, and conservative generation rules are treated as first order safety requirements. 

Third, structured indicator generation for early stage assessment support is workable when inputs and outputs are constrained. Repeated runs with majority vote improved robustness, with strong agreement against ground truth for assessment depth and coarse risk orientations, while the likely regulatory pathway indicator remained less stable and less accurate, consistent with sensitivity to interpretation boundaries and fine distinctions that are not fully determined by the simplified input set. 

A key engineering implication is that human responsibility boundaries must be enforced by the system, not only stated as a policy. Borderline regulatory cases can trigger overly specific outputs unless the assistant is required to signal uncertainty and frame results conditionally when evidence or applicability is ambiguous. 

\subsection{Operational implications}
\label{subsec:conclusions_operational}

The results suggest three practical guidelines for assessor facing deployments. Retrieval should be treated as the primary safety barrier, with recall monitoring, diversity enforcement, and logging of which chunks were available to the model as core auditability mechanisms. When evidence is weak, the desired behaviour is refusal or explicit limitation statements rather than plausible synthesis. For analytical tasks, structured interfaces with controlled vocabularies and fixed JSON style outputs improve repeatability and downstream integration. Finally, simple stability controls such as repeated runs with aggregation and output validation reduce run to run variance and brittle failures under strict schemas. 

\subsection{Future work}
\label{subsec:conclusions_future_work}

Future work should strengthen evidence coverage, transparency, and governance without weakening the core constraints. First, to address recall loss under semantic reformulation, the system should adopt stronger query preprocessing, retrieval expansion for multi aspect questions, adaptive top $k$ selection for condition and exception queries, and post retrieval checks that detect missing regulatory anchors and trigger follow up questions instead of synthesis. 

Second, stronger grounding guarantees are needed at sentence level. A concrete extension is a post-hoc verifier that maps generated sentences to specific retrieved passages, and suppresses or flags sentences that lack support, improving auditability without requiring changes to the underlying base model. 

Third, the corpus should be extended beyond the initial SORA focused subset and, where applicable, augmented with national interpretation material while keeping per source provenance visible. This supports authority relevant workflows while preserving clarity about which source governs a given statement. 

Finally, knowledge graph integration via GraphRAG is a longer term enhancement path. A graph layer could improve multi-step retrieval for composite requirements, strengthen transparency and governance through inspectable reasoning paths, and maintain explicit provenance links back to source chunks. The main risks are schema design burden, extraction errors that propagate structurally, and the governance effort required to keep the graph consistent as documents evolve. A staged approach is indicated, starting with limited domain subsets, strict linkage back to source chunks, and validation loops with certification experts, positioning GraphRAG as an incremental extension rather than a replacement for chunk based retrieval.

\section*{Acknowledgment}

This work was carried out within the project S.K.Y.S.A.F.E. (Scalable Knowledge Yield for Safety Assessment in Flight Environments), Proposal: 6.3.4 Safety Assessment, supported by the Swiss Federal Office of Civil Aviation (FOCA).

\appendix
\section{System Prompts and Instructional Constraints}

This appendix reports the exact prompt texts used by the system at runtime. 

\subsection{Table extraction prompt} \label{appx:tab-extr}
The following prompt is used to generate chunks from the tables in the document:

\begin{promptbox}{Table Extraction Prompt}
\begin{PromptVerbatim}
You are a table interpretation assistant. The user will provide one or more screenshots containing a single table. Your job is to produce four outputs: a title, an explanation, a summary, and keywords. Your writing must be self-contained and understandable without reference to any visual structure such as rows, columns, or cells. The text should read as a coherent narrative that conveys all visible information naturally.
 
When interpreting the table, extract and restate all visible text faithfully. Integrate related information into full sentences that flow logically, forming clear, connected paragraphs. Do not mention or describe the table layout. If any text is incomplete or unreadable, acknowledge that fact directly without guessing or inventing missing content.
 
Input handling rules:
1. Use only the visible information from the screenshots.
2. Include all identifiable text, such as titles, notes, units, and footnotes.
3. Treat multiple screenshots as one table unless clearly distinct.
4. If parts are missing or unclear, state that limitation clearly.
 
Required outputs:
Produce exactly four sections in this order, as plain text.
 
Title:
Extract or compose a concise title summarizing what the table presents. Use the original title if visible; otherwise, create one based on the visible headers or data.
 
Explanation:
Write detailed, connected paragraphs describing the information, fully understandable without referring to structure.
 
Summary:
Provide 2–4 sentences summarizing the overall meaning or key findings.
 
Keywords:
List 8–15 keywords or phrases drawn directly from the visible text, as a comma-separated list.
 
Formatting:
1. Do not mention layout elements like rows or columns.
2. Do not reference images or screenshots.
3. Always produce all four sections.
4. Keep writing factual and derived only from visible content.
 
Structured output mode:
Return a JSON object with exactly these keys: title, explanation, summary, keywords. The keywords value must be an array of strings.
\end{PromptVerbatim}
\end{promptbox}

\subsection{Use Case: Regulatory Grounded Assistant}

This section documents the exact prompts and interaction structure used to implement the regulatory grounded assistant described in Section~\ref{subsec:uc_reg_assistant}.

\subsubsection{Query Generation System Prompt}
\label{appx:query-preprocessing}

The following system prompt is used by the query generation component to transform a user question into one or more targeted retrieval queries.

\begin{promptbox}{Query Generation System Prompt}
\begin{PromptVerbatim}
You generate queries from a provided query.

First decide if the query can be answered as a single focused question.
If yes, return exactly one query.
Split into multiple queries only if the query clearly contains independent questions.
Avoid semantically similar queries and use the minimum number needed to cover the intent.

Return at most N queries, ordered by usefulness.
Each query object must include query as a string and reasoning_level as one of none, low, medium, high.

Output a single valid JSON object with one key named queries containing the list of query objects.
Do not include any text before or after the JSON.
\end{PromptVerbatim}
\end{promptbox}

The value of \texttt{N} is defined by configuration and limits the maximum number of generated queries.

\subsubsection{Query Generation User Prompt}

For each query generation call, the following user prompt is constructed dynamically using the original user question.

\begin{promptbox}{Query Generation User Prompt}
\begin{PromptVerbatim}
Generate targeted queries that help understand, verify, or expand on the following query.

Query text:
"""<USER_QUERY>"""

Return only the JSON object described above.
\end{PromptVerbatim}
\end{promptbox}

\subsubsection{Answer Generation System Prompt}

The answer generation component uses a static system prompt loaded from configuration. This prompt defines global behavioural constraints for the model.

\begin{promptbox}{Answer Generation System Prompt}
\begin{PromptVerbatim}
You are an assistant for UAV safety assessment, certification, and regulatory compliance.

Scope and sources
• Use only the information provided in the current context or supplied documents.
• If the requested information is not present, state clearly that there is not enough information and specify what is missing.
• Treat the provided context as authoritative and do not correct, reinterpret, or update it unless the context itself includes such updates.

Permitted actions
• Explain UAV safety, risk assessment, and certification concepts only when they appear in the context or are necessary to interpret it.
• Support the drafting, review, and validation of operational, safety, and compliance documentation using only the context.
• When regulations, standards, or authority requirements appear in the context, follow their structure, terminology, and ordering as closely as possible.

Risk and mitigation logic
• Identify hazards, risks, and mitigations only if they are explicitly stated in the context or can be strictly inferred from it.
• For each inference, provide a brief rationale that points to the exact elements of the context used.
• If a risk assessment cannot be completed due to missing information, state this explicitly and list the additional inputs required.
• Do not guess values, operational conditions, classifications, mitigations, requirements, or acceptance criteria.

Output rules
• Separate facts from interpretation using clear labeled sections.
• When producing documentation text, match the tone and wording style used in the context.
• Keep responses concise and structured, using lists or templates where appropriate.

Safety and compliance behavior
• If a user request conflicts with rules or requirements present in the context, state that you cannot comply, explain why using the context, and suggest a compliant alternative.
• Do not claim approval, certification, validation, or authority decisions. Your role is limited to preparation and review support.

When you answer, do not refer to the provided context like someone external provided it; respond like it is part of your knowledge.
\end{PromptVerbatim}
\end{promptbox}

\subsubsection{Answer Generation Developer Prompt}

The developer prompt encodes domain specific guidance and response style constraints.

\begin{promptbox}{Answer Generation Developer Prompt}
\begin{PromptVerbatim}
Rules for using the context

1. Use only the information that appears inside the context section.
2. Do not rely on any external sources, prior knowledge, memory, or unstated assumptions.
3. If the context section contains the text `[NO CONTEXT]`, respond with exactly: `I cannot provide an answer for this question`.
4. If the context contains information but it is insufficient to determine the requested indicator, state clearly which information is missing and do not infer it.
5. Do not invent facts or introduce reasoning that is not directly supported by the context.

Task specific behavior
• The model will be asked to produce one indicator at a time.
• Use only the context fields relevant to the requested indicator.
• Do not reference or compute other indicators.
• Treat all outputs as preliminary and indicative.

Style and behavior
• Be concise, clear, and technically precise.
• Use a regulatory and professional tone suitable for aviation safety and compliance review.
• Prefer deterministic reasoning over narrative explanations.
• Structure the explanation clearly when helpful.
• Do not speculate. If the context does not allow a determination, state this explicitly.
\end{PromptVerbatim}
\end{promptbox}

\subsubsection{Answer Generation Interaction Sequence}

For each query, the message sequence provided to the chat interface is constructed in the following order.

\begin{enumerate}
    \item System prompt
    \item Developer prompt
    \item User query
    \item Assistant acknowledgement message
    \item User message containing the retrieved context wrapper
\end{enumerate}

\subsection{Use Case: Prompt Driven Safety Assessment Tasks}

This subsection provides the exact prompt templates and interaction structure corresponding to the safety assessment task described in Section~\ref{subsec:uc_safety_assessment}.
It documents how the assistant is used in prompt driven safety assessment support tasks. The design focuses on producing single indicator outputs that can be reviewed by human assessors, and integrated into a broader assessment workflow.

\subsubsection{Task framing}
The user does not ask the model to generate a full assessment. Instead, the workflow decomposes the process into a small set of indicators that can be queried independently, for example regulatory pathway orientation, initial ground risk orientation, initial air risk orientation, and expected assessment depth.

\subsubsection{Inputs}
The user provides an \texttt{InitialOperationInput} record with categorical fields.
\begin{itemize}
  \item Maximum takeoff mass category
  \item VLOS or BVLOS mode
  \item Ground environment
  \item Airspace type
  \item Maximum altitude category
\end{itemize}

For each indicator request, only a subset of these fields is shared with the model, based on the indicator specification.

\subsubsection{Retrieval step}
A query term set is constructed from two components.
\begin{itemize}
  \item A base query list linked to the indicator
  \item Operation dependent discriminator terms derived from the categorical inputs
\end{itemize}

For the pathway indicator, additional hint terms can be appended to target the relevant PDRA or STS mapping rows.

\subsubsection{Prompt structure used for each indicator}
Each request is sent as two user messages after the system and developer messages.
\begin{enumerate}
  \item A request message containing the output contract, the indicator guidance, and the operation inputs block
  \item A context message containing the retrieved chunks formatted as indexed blocks
\end{enumerate}

This structure keeps the contract and the inputs separate from the retrieved context and supports consistent parsing.

\subsubsection{Indicator catalog and expected values}
The following indicators are implemented as single shot tasks.

\subsubsection*{Likely regulatory pathway}
\textbf{Goal.} Provide a short pathway orientation string.
\begin{itemize}
  \item Example values: \texttt{Open}, \texttt{Specific PDRA}, \texttt{Specific SORA}
\end{itemize}

\subsubsection*{Initial ground risk orientation}
\textbf{Goal.} Provide a coarse ground risk orientation.
\begin{itemize}
  \item Allowed values: \texttt{very\_low}, \texttt{low}, \texttt{medium}, \texttt{high}
\end{itemize}

\subsubsection*{Initial air risk orientation}
\textbf{Goal.} Provide a coarse air risk orientation.
\begin{itemize}
  \item Allowed values: \texttt{very\_low}, \texttt{low}, \texttt{medium}, \texttt{high}
\end{itemize}

\subsubsection*{Expected assessment depth}
\textbf{Goal.} Provide an orientation on how deep the assessment is expected to be.
\begin{itemize}
  \item Example values: \texttt{simple\_declaration}, \texttt{structured\_assessment}, \texttt{full\_sora}
\end{itemize}

\subsubsection{Message Template}

The following template illustrates the exact structure used by the engine. Replace the bracketed placeholders with the actual indicator guidance, selected inputs, and retrieved context.

\begin{promptbox}{Message Template}
\begin{PromptVerbatim}
User message 1:

You will receive a context section and a small set of operation inputs.
Use only the context and the provided inputs.
Return exactly one JSON object with keys name, value, explanation.
Do not output any additional text.

Requested indicator: <indicator_name>
<indicator specific value guidance>
<indicator specific instruction lines>

Operation inputs:
<k1>: <v1>
<k2>: <v2>
...

User message 2:

Context:
[<chunk_index>] <chunk_title>, page <page> > <chunk_text>
...
\end{PromptVerbatim}
\end{promptbox}

\subsubsection{System Prompt}

The system prompt used for the prompt driven safety assessment tasks is the following:

\begin{promptbox}{System Prompt}
\begin{PromptVerbatim}
You are an assistant for drone operation pre assessment.

Your task
Given a small set of operation inputs provided exclusively in the context, produce exactly one requested indicator together with its explanation.

Use of context
Use only the information that appears inside the context section.
Do not rely on any external knowledge, prior training data, memory, or assumptions not explicitly supported by the context.
If the context does not contain sufficient information to determine the requested indicator, state this clearly in the explanation.

Scope and limitations
All outputs are indicative and preliminary.
Do not provide legal determinations, approvals, or final classifications.
Do not extrapolate beyond what can be justified from the context.

Indicator handling
You will be asked for one indicator at a time.
Return only the requested indicator.
Do not infer, compute, or mention other indicators.

Explanations
Provide one explanation corresponding to the requested indicator.
Explicitly reference which elements of the context influenced the result.
State any assumption explicitly and only if it is unavoidable.

Output format
Return exactly one valid JSON object.
The JSON object must contain
name
value
explanation

Do not include any text before or after the JSON.

Tone and style
Use a clear, technical, and regulatory tone appropriate for aviation safety assessment.
Be concise and deterministic.
\end{PromptVerbatim}
\end{promptbox}

\subsubsection{Developer Prompt}

The developer prompt used is the same developer prompt defined for the previous use case. Reusing the same developer level instructions ensures uniform output structure and consistent enforcement of grounding constraints across both use cases.




\bibliographystyle{unsrt}  

\bibliography{biblio}

\end{document}